\documentclass[runningheads,a4paper]{llncs}

\usepackage{amssymb}
\usepackage{amsmath}
\usepackage{graphicx}
\usepackage{booktabs}
\usepackage{rotating}

\usepackage{url}
\urldef{\mailsa}\path|{alfred.hofmann, ursula.barth, ingrid.haas, frank.holzwarth,|
\urldef{\mailsb}\path|anna.kramer, leonie.kunz, christine.reiss, nicole.sator,|
\urldef{\mailsc}\path|erika.siebert-cole, peter.strasser, lncs}@springer.com|    
\newcommand{\keywords}[1]{\par\addvspace\baselineskip
\noindent\keywordname\enspace\ignorespaces#1}

\newcommand\ci{\perp\!\!\!\perp}

\begin{document}

\mainmatter  

\title{Individualized Risk Prognosis for Critical Care Patients: A Multi-task Gaussian Process Model}

\titlerunning{Individualized Risk Prognosis for Critical Care Patients} 

%
%
\author{Ahmed M. Alaa\inst{1} \and Jinsung Yoon\inst{1} \and
Scott Hu\inst{2} \and Mihaela van der Schaar\inst{3,4,1}}
\authorrunning{A. M. Alaa et al.} 
%
\tocauthor{Ahmed M. Alaa, Jinsung Yoon, Scott Hu, and Mihaela van der Schaar}
\institute{Electrical Engineering Department, University of California, Los Angeles, USA,\\
\email{\{ahmedmalaa,jsyoon0823\}@ucla.edu},
\and
David Geffen School of Medicine, University of California, Los Angeles, USA,\\
\email{scotthu@mednet.ucla.edu},
\and
Department of Engineering Science, University of Oxford, UK,\\
\email{mihaela.vanderschaar@eng.ox.ac.uk},
\and 
Alan Turing Institute, London, UK}

\maketitle 

\begin{abstract}
We report the development and validation of a {\it data-driven} real-time risk score that provides timely assessments for the clinical acuity of ward patients based on their temporal lab tests and vital signs, which allows for timely intensive care unit (ICU) admissions. Unlike the existing risk scoring technologies, the proposed score is ``individualized"-- it uses the {\it electronic health record} (EHR) data to cluster the patients based on their static covariates into {\it subcohorts} of ``similar" patients, and then learns a separate temporal, non-stationary multi-task {\it Gaussian Process} (GP) model that captures the physiology of every subcohort. Experiments conducted on data from a heterogeneous cohort of 6,094 patients admitted to the Ronald Reagan UCLA medical center show that our risk score significantly outperforms the state-of-the-art risk scoring technologies, such as the {\it Rothman index} and {\it MEWS}, in terms of timeliness, true positive rate (TPR), and positive predictive value (PPV). In particular, the proposed score increases the AUC with 20$\%$ and 38$\%$ as compared to Rothman index and MEWS respectively, and can predict ICU admissions 8 hours before clinicians at a PPV of 35$\%$ and a TPR of 50$\%$. Moreover, we show that the proposed risk score allows for better decisions on when to discharge clinically stable patients from the ward, thereby improving the efficiency of hospital resource utilization.
\end{abstract}

\section{Introduction}
Prognostic risk assessment models that quantify the acuity of critical care patients in real-time can inform vital and delay-critical clinical decision-making \cite{churpek2014using}. Unanticipated adverse events such as mortality, cardiopulmonary arrest, or intensive care unit (ICU) transfer are often preceded by disorders in a patient's physiological parameters \cite{kause2004comparison}. Timely prediction of such events can be carried out by continuously quantifying the patient's acuity using evidence in her physiological parameters, and hence assessing her risk for a specific event by prompting a real-time ``risk score" that can be tracked by clinicians. 

Current clinical practice in most hospitals and healthcare facilities rely on two categories of risk scoring technologies. The first category comprises {\it early-warning systems} (EWS), such as MEWS \cite{morgan1997early}, which hinge on {\it expert-based} models for triggering transfer to the ICU. A major drawback of ``expert-based" scores is that they are not subject to any rigorous, objective validation. Recent systematic reviews have shown that EWS-based scores only marginally improve patient outcomes while substantially increasing clinician and nursing workloads, leading to alarm fatigue and inefficient resource utilization \cite{tsien1997poor,cvach2012monitor,bliss2000behavioural,subbe2001validation}. The second category of risk scores are based on data-intensive regression models that are built using the electronic health record (EHR) data. The most notable technology in this category is the Rothman index \cite{rothman2013development}, which is currently deployed in more than 70 hospitals in the US (including the Houston Methodist hospital in Texas, and Yale-New Haven hospital in Connecticut) \cite{WSJ}, and was recently shown to be superior to MEWS-based models in terms of false alarm rates \cite{finlay2014measuring}. 

While the Rothman index offers a significant performance improvement over MEWS, it suffers from 2 major drawbacks. First, it adopts a ``one-size-fits-all" risk scoring model that ignores the individual traits of the monitored patients. Second, it ignores the temporal aspect of the physiological data as it computes the patient's risk score at a particular moment using the patient's vital signs at that moment, ignoring her temporal physiological trajectory (see Figure 1 in \cite{rothman2013development}). To that end, we report the development and validation of a novel data-driven real-time risk score that addresses these drawbacks and provides a significant performance improvement over all existing clinical risk scoring technologies. The proposed risk score is a numeric value between 0 and 1 that corresponds to the patient's risk for clinical deterioration, and is computed and updated in real-time by aggregating two types of a monitored patient's information:  
\begin{enumerate}
\item {\bf Static Admission Information:} this includes all the static information gathered about the patient upon her hospitalization and remain fixed during her stay in the ward (e.g. age, gender, race, ICD-9 code, diagnosis, etc). \\
\item {\bf Time-Varying Physiological Information:} this includes all the physiological parameters (vital signs and lab tests) that are repeatedly gathered for the patient during her stay in the ward (e.g. systolic and diastolic blood pressure, $O_2$ saturation, respiratory rate, Glucose, Glasgow coma scale score, Creatinine, etc).  
\end{enumerate}
Our risk score models a patient's entire temporal physiological trajectory via a non-stationary multi-task {\it Gaussian Process} (GP) model, which captures irregularly sampled and temporally correlated physiological data \cite{ghassemi2015multivariate}. The model parameters are learned in a data-driven fashion: we use the EHR data in order to fit the multi-task GP {\it hyper-parameters} for clinically stable patients (patients who were recorded in the EHR as discharged from the ward), and fit a different set of hyper-parameters for clinically deteriorating patients (patients who were recorded in the EHR as transferred to the ICU). The patient's risk score is computed as the optimal test statistic of a {\it sequential hypothesis test} that tests the hypothesis that the patient is clinically deteriorating given a sequence of physiological measurements \cite{wald1973sequential}. Following the newly emerging concepts of {\it precision medicine}, we ensure that our risk scoring procedure is tailored to the individual's traits by introducing {\it latent phenotype} variables, where a phenotype represents a distinct way in which a patient manifests her clinical status \cite{saria2015subtyping}. Using unsupervised learning, we discover the number of patient phenotypes from the EHR data, learn the association between a patient's static admission information with her phenotype, and calibrate a separate model for every phenotype.

\begin{table}[t]
    \centering
    \caption{Physiological data and admission information associated with each patient in the cohort under study.}
\begin{tabular}{ccc}
\toprule
\multicolumn{2}{c}{\small {\bf Time-Varying Physiological Data}} \\
\cmidrule(r){1-2}
{\bf \small Vital Signs} & {\bf \small Lab Tests} & {\bf \small Static Admission Information} \\
\midrule
\small Diastolic blood pressure      & \small Glucose                & \small Age                   \\
\small Eye opening                   & \small Urea Nitrogen          & \small Admission floor       \\
\small Glasgow coma score            & \small White blood cell       & \small Gender                \\
\small Heart rate                    & \small Creatinine             & \small Stem cell transplant  \\
\small Respiratory rate              & \small Hemoglobin             & \small ICD-9 code            \\
\small Temperature                   & \small Platelet Count         & \small Transfer status       \\
\small $O_2$ Device Assistance       & \small Potassium              &        \\
\small $O_2$ Saturation              & \small Sodium                 &        \\
\small Best motor response           & \small Total $CO_2$           &        \\
\small Best verbal response          & \small Chloride               &        \\
\small Systolic blood pressure       &                               &        \\
\bottomrule
\end{tabular}
\label{Tab1}
\end{table}

\section{Study Subjects}
We conducted our experiments using an EHR dataset for a heterogeneous cohort of 6,094 patients admitted to the Ronald Reagan UCLA medical center in a period that spans 3 years (March 2013 to March 2016). The patients' cohort is quite heterogeneous; we considered admissions to all units in the medical center, including the cardiac observation unit, cardiothoracic unit, hematology and stem cell transplant unit and the liver transplant service. The cohort comprised patients with a wide variety of ICD-9 codes and medical conditions, including leukemia, hypertension, sepsis, abdomen, pneumonia, and renal failure.

Every patient record in the dataset is associated with the time-varying and static information listed in Table \ref{Tab1}. The time-varying physiological measurements are collected over irregularly spaced time intervals (usually ranging from 1 to 4 hours); for each physiological time series, we have access to the times at which each value was gathered. The patients' length of stay in the wards ranged from 4 hours to 2700 hours. Every patient in the cohort is associated with a label that indicates whether the patient was discharged home or transferred to the ICU, and hence we know which patients were clinically stable and which ones were clinically deteriorating. Around 9$\%$ of the patients in the cohort were clinically deteriorating and experienced an unanticipated ICU transfer. We excluded all patients for whom a routine ICU transfer was preordained since the objective of the proposed risk score is to predict unanticipated ICU admissions.

\section{The Proposed Individualized Risk Score}  
\subsection{Notations and Definitions}
\label{Sec3.0}
Let $X_i(t) = [X_{i1}(t), X_{i2}(t), .\,.\,., X_{iD}(t)]^{T}$ be a $D$-dimensional stochastic process representing the $D$ time-varying physiological streams for patient $i$. In our dataset, $D = 21$, i.e. the number of lab tests and vital signs listed in Table \ref{Tab1}. The vital signs and lab tests of patient $i$ are gathered at arbitrary time instances $\{t^{i}_{dj}\}_{d=1,j=1}^{D, M_{d}}$ (where $t=0$ is the time at which the patient is admitted to the ward), where $M_d$ is the total number of measurements of the $d^{th}$ vital sign (or lab test) that where gathered during the patient's stay in the ward. Thus, the set of all observations of the physiological data that clinicians collect for a specific patient during her stay in the ward is given by $\{X_i(t^{i}_{dj})\}_{d=1,j=1}^{D, M_{d}}$, and we will refer to the realizations of these variables as $\{x^i_{dj}, t^i_{dj}\}_{i,d,j}$.

We define the $S$-dimensional random vector $Y_i$ as patient $i$'s static admission information, i.e. in our dataset, $S=6$ as indicated in Table \ref{Tab1}. We denote the realizations of patient $i$'s static information as $Y_i = y_i$. Thus, the set of all information associated with a patient can be gathered in a set $\{y_i, x^i_{dj}, t^i_{dj}\}_{i,d,j}$.
  
\subsection{Risk Scoring as a Sequential Hypothesis Test}
\label{Sec3.1}
Let $V_i \in \{0,1\}$ be a binary {\it latent} variable that corresponds to patient $i$'s true clinical status, where $V_i=0$ stands for a stable clinical status, and $V_i=1$ stands for a deteriorating clinical status. We assume that $V_i$ is a fixed {\it latent class} that determines the physiological model, i.e. $V_i$ is drawn randomly for patient $i$ at admission time and stays fixed over the patient's stay in the ward. In the EHR dataset, the value of $V_i$ is revealed at the end of every physiological stream, where $V_i=1$ if the patient is admitted to the ICU, and $V_i=0$ if the patient is discharged home. 

Since physiological streams manifest the patients' clinical statuses, it is natural to assume that the conditional distributions of $\{x^i_{dj}, t^i_{dj}\}_{i,d,j}$ given $V_i=0$ differs from that of $\{x^i_{dj}, t^i_{dj}\}_{i,d,j}$ given $V_i=1$. Our conception of the risk score can be described as follows. During the patient's stay in the ward, we are confronted with two hypotheses: the null hypothesis $\mathcal{H}_{o}$ corresponds to the hypothesis that the patient is clinically stable, whereas the alternative hypothesis $\mathcal{H}_{1}$ corresponds to the hypothesis that the patient is clinically deteriorating, i.e.   
\begin{equation}
V_i = 
\left\{
\begin{array}{ll}
      0: \,\, \mathcal{H}_{o} \,\, (\mbox{clinically stable patient}),\\
      1: \,\, \mathcal{H}_{1} \,\, (\mbox{clinically deteriorating patient}).\\
\end{array} 
\right.
\label{eqq1} 
\end{equation}
Thus, the prognosis problem is equivalent to a {\it sequential hypothesis test} \cite{wald1973sequential}, i.e. the clinicians need to reject one of the hypotheses at some point of time after observing a series of physiological measurements. Hence, we view the patient's risk score as the test statistic of a sequential hypothesis test: patient $i$'s risk score at time $t$, which we denote as $R_i(t)\in [0,1]$, is the posterior probability of hypothesis $\mathcal{H}_{1}$ given the observations $\{y_i, x^i_{dj}, t^i_{dj} \leq t\}_{i,d,j}$, i.e. $R_i(t) = \mathbb{P}\left(\mathcal{H}_{1}\left|\{y_i, x^i_{dj}, t^i_{dj} \leq t\}_{i,d,j}\right.\right)$. Using Bayes' rule we have that  
\begin{align}
R_i(t) = \frac{\mathbb{P}\left(\left. \{y_i, x^i_{dj}, t^i_{dj} \leq t\}_{i,d,j}\right|\mathcal{H}_{1}\right) \cdot \mathbb{P}\left(\mathcal{H}_{1}\right)}{\sum_{v \in \{0,1\}}\mathbb{P}\left(\left. \{y_i, x^i_{dj}, t^i_{dj} \leq t\}_{i,d,j}\right|\mathcal{H}_{v}\right) \cdot \mathbb{P}\left(\mathcal{H}_{v}\right)},
\label{eqq2} 
\end{align}
where $\mathbb{P}\left(\mathcal{H}_{1}\right) = \mathbb{P}\left(V_i=1\right)$ is the prior probability of a patient in the ward being admitted to the ICU (i.e. the rate of ICU admissions, which is 9$\%$ in our dataset). In order to be able to compute the risk score in (\ref{eqq2}), we specify the conditional distributions $\mathbb{P}\left(\left. \{y_i, x^i_{dj}, t^i_{dj} \leq t\}_{i,d,j}\right|\mathcal{H}_{v}\right), v \in \{0,1\}$ in the next Subsection.

\subsection{The Non-stationary Multi-task GP Model}
\label{Sec3.2}
We adopt a multi-task GP model as the distributional specification for the continuous-time process $X_i(t)$; our choice for a multi-task GP is motivated by its ability to capture irregularly sampled, multi-variate time series data \cite{ghassemi2015multivariate}. We specify a different set of GP hyper-parameters for the physiological data generated under $V_i=0$ and those generated under $V_i=1$, i.e.     
\begin{equation}
X_i(t)|V_i=v \sim \mathcal{GP}({\bf \Theta}_{v}),
\label{eqq3} 
\end{equation}
where ${\bf \Theta}_{v}$ is the hyper-parameter set of the GP model for patients with $V_i=v$.

Since hospitalized patients are subject to external clinical interventions, and are likely to exhibit a progression of subsequent ``phases" of clinical stability or deterioration, a stationary covariance kernel, such as the one defined in \cite{ghassemi2015multivariate}, would do not suffice to describe the patients' entire physiological trajectory. This motivates a non-stationary model for $X(t)$ in which the time domain is divided into a sequence of $K$ {\it epochs}, where the $k^{th}$ epoch for patient $i$ has a duration of $T^i_k$ and is described by a {\it locally stationary} covariance kernel. This is achieved by assigning a different set of mean and stationary covariance hyper-parameters ${\bf \Theta}_{v,k}$ for every epoch $1 \leq k \leq K$, and assuming that the physiological data in different epochs are independent. The duration of epoch $k$ for patient $i$, $T^i_{k}$, is an integer number of hours that is drawn from a {\it negative binomial distribution} $f_{kv}(T|\lambda_{kv})$ with a parameter $\lambda_{kv}$. Since patients arrive at the hospital ward at random time instances, at which the clinical status is unknown, we define $\bar{k}_i \in \{1,2,.\,.\,.,K\}$ as a latent initial epoch index for patient $i$, which we assume to be drawn from a {\it multinomial} distribution. Thus, the clinicians observe physiological measurements drawn from a process with the underlying epoch index sequence $\{\bar{k}_i, \bar{k}_i+1,.\,.\,., K\}$, with random epoch durations $\{T^i_{\bar{k}_i}, T^i_{\bar{k}_i+1},.\,.\,., T^i_{K}\}$, and for which the GP epoch-specific hyper-parameters are $\{{\bf \Theta}_{v,\bar{k}_i}, {\bf \Theta}_{v,\bar{k}_i+1},.\,.\,., {\bf \Theta}_{v,K}\}$. Note that we assume that all the patients' epoch sequences end with epoch $K$, this ensures that all the physiological time series of all patients in the cohort are temporally {\it aligned}, which is essential for proper learning of the model parameters. 

The GP hyper-parameters ${\bf \Theta}_{v,k}$ for clinical state $v$ and epoch $k$ comprise a constant mean functions $m_{v,k}$ and a {\it squared exponential} covariance kernel with an {\it intrinsic correlation model} for the correlations between the different vital signs and lab tests \cite{bonilla2007multi}. That is, the covariance kernel $K_{v,k}(u,w,t,t^{\prime})$ which quantifies the correlation between the physiological measurements $X_{iu}(t)$ and $X_{iw}(t^{\prime})$ is given by
\begin{align}
K_{v,k}(u,w,t,t^{\prime}) = {\bf \Sigma}_{v,k}(u,w)\,k_{v,k}(t,t^{\prime}),
\label{eqq4}
\end{align}
if $t$ and $t^{\prime}$ belong to the same epoch, and $k(u,v,t,t^{\prime}) = 0$ otherwise. In (\ref{eqq4}), ${\bf \Sigma}_{v,k}$ is a positive semi-definite correlation matrix, and $k_{v,k}(t,t^{\prime})$ is a squared exponential covariance kernel given as follows 
\begin{equation}
k_{v,k}(t,t^{\prime}) = \mbox{exp}\left(-\frac{1}{2\ell^{2}_{v,k}}\,||t-t^{\prime}||^{2}\right),
\label{eqq5}
\end{equation} 
where $\ell_{v,k}$ is the {\it characteristic length scale} parameter of the GP. We denote the set of all GP hyper-parameters under clinical status $v$ as ${\bf \Theta}_{v} = \{{\bf \Theta}_{v,1},.\,.\,.,{\bf \Theta}_{v,K}\}$, and the epoch duration parameters as ${\bf \Lambda}_{v} = \{{\bf \lambda}_{v,1},.\,.\,.,{\bf \lambda}_{v,K}\}$. 

\begin{figure*}[t!]
    \centering
    \includegraphics[width=2.5 in]{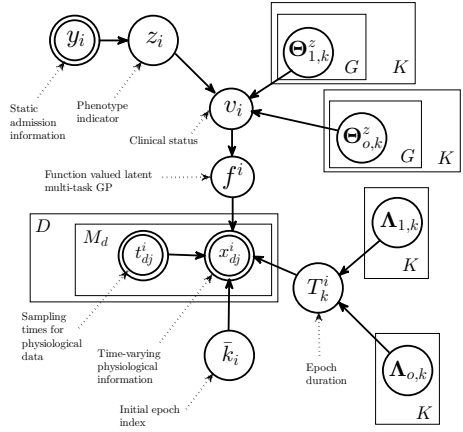}
		\caption{\small Graphical depiction for the proposed physiological model.}
		\label{Fiq3}
\end{figure*}

\subsection{The Latent Phenotype Indicators}
\label{Sec3.3}
The model presented in Subsection \ref{Sec3.2} is a ``one-size-fits-all" model that treats all patients in the same way since it does not incorporate the patients' baseline static features; however, in reality, different patients manifest their clinical status in different ways depending on their traits-- the different ways in which patients manifests their physiology are known as the {\it phenotypes} \cite{saria2015subtyping}. In this Subsection, we refine the model in Subsection \ref{Sec3.2} to ensure {\it individualization}, i.e. ensure that our physiological model is tailored to the individual traits, by introducing a {\it latent phenotype indicator variable} $Z_i \in \{1,2,.\,.\,.,G\}$ as the phenotype to which patient $i$ belongs. We assume that the phenotype indicator variable $Z_i$ possesses the following properties: $Z_i \ci V_i \,|\, Y_i,$ and $V_i \ci Y_i \,|\, Z_i$.

We assume that a separate GP model is associated with every phenotype, i.e. for phenotype $z \in \{1,2,.\,.\,.,G\}$, the corresponding GP hyper-parameter set is ${\bf \Theta}^z_{v}$, and the epoch duration parameter set is ${\bf \Lambda}^z_{v}$. The phenotype indicators are latent and hence we do not know to which phenotype patient $i$ belongs upon admission. However, since a patient's phenotype naturally depends on her age, race, diagnosis, etc, we can infer the patient's phenotype by estimating the posterior probability of patient $i$'s membership in phenotype $z$, which we denote as $\gamma_{z}(Y_i) = \mathbb{P}(Z_i = z\,|\,Y_i)$. Thus, the resulting physiological model is a {\it mixture model} that combines $G$ instantiations of the model in Subsection \ref{Sec3.2} with weights that are proportional to the phenotype memberships $\{\gamma_{1}(Y_i),.\,.\,.,\gamma_{G}(Y_i)\}$. The model parameters ${\bf \Theta}^z_{v}$ and ${\bf \Lambda}^z_{v}$ are estimated from the dataset using the standard expectation-maximization (EM) algorithm, and we use the {\it Bayesian information criterion} to select the number of phenotypes $G$ and the number of epochs $K$.  

Figure \ref{Fiq3} depicts a graphical model for the patients' physiological data. In our previous works in \cite{yoon2016forecasticu} and \cite{alaa2016semi}, we developed {\it ForecastICU}, a GP-based risk score that we have shown to be superior to both the Rothman index and MEWS. ForecastICU is a subset of the proposed model that does not consider individualization and non-stationarity; we compare the performance of the proposed risk model with ForecastICU in the next Section.         

\section{Results}
\begin{figure}[!tbp]
  \centering
  \begin{minipage}[b]{0.4\textwidth}
\includegraphics[width=2.5in]{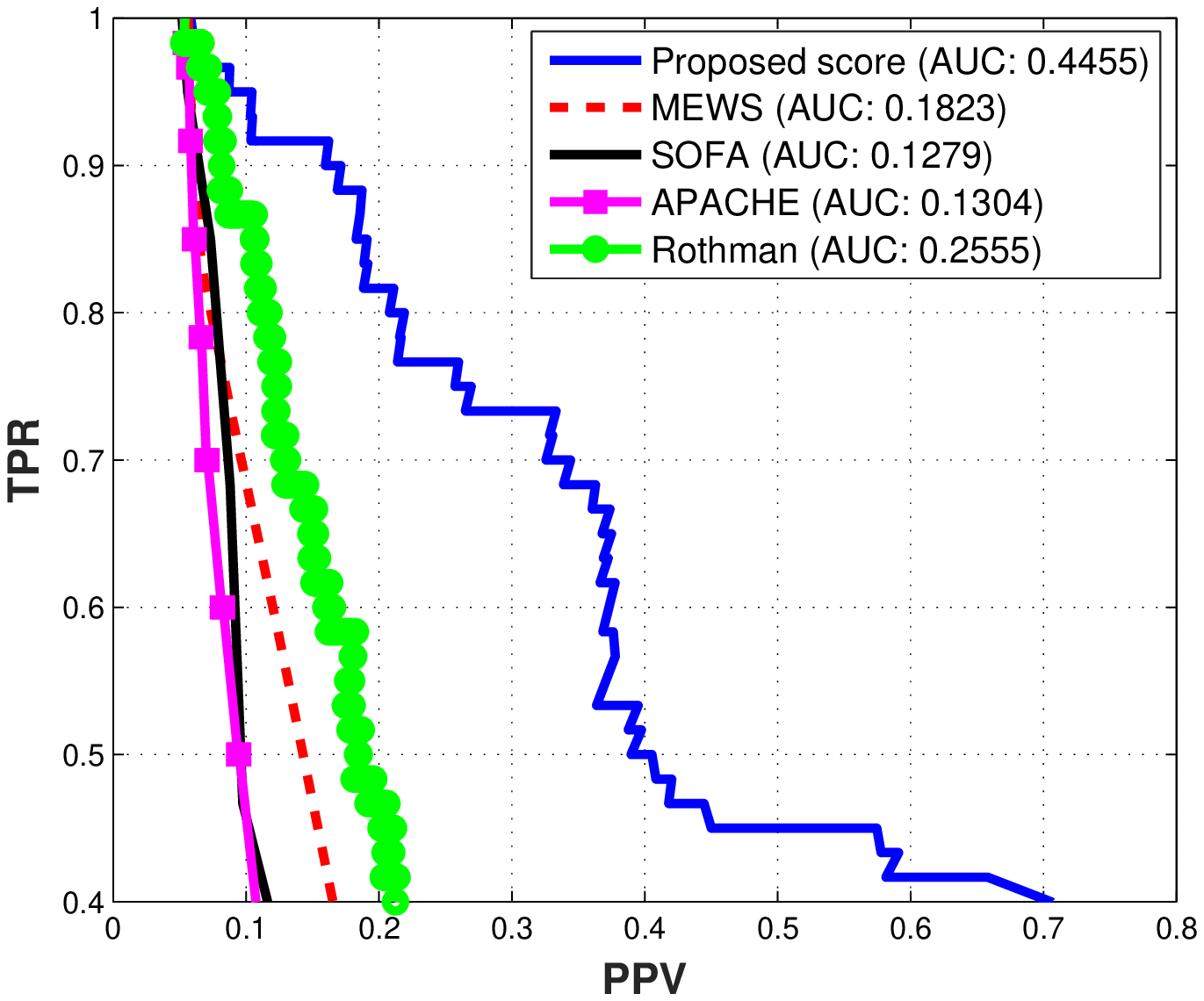}
        \caption{\small ROC curve (TPR vs PPV).}
	\label{Figa1}			
  \end{minipage}
  \hfill
  \begin{minipage}[b]{0.5\textwidth}
\includegraphics[width=2.5in]{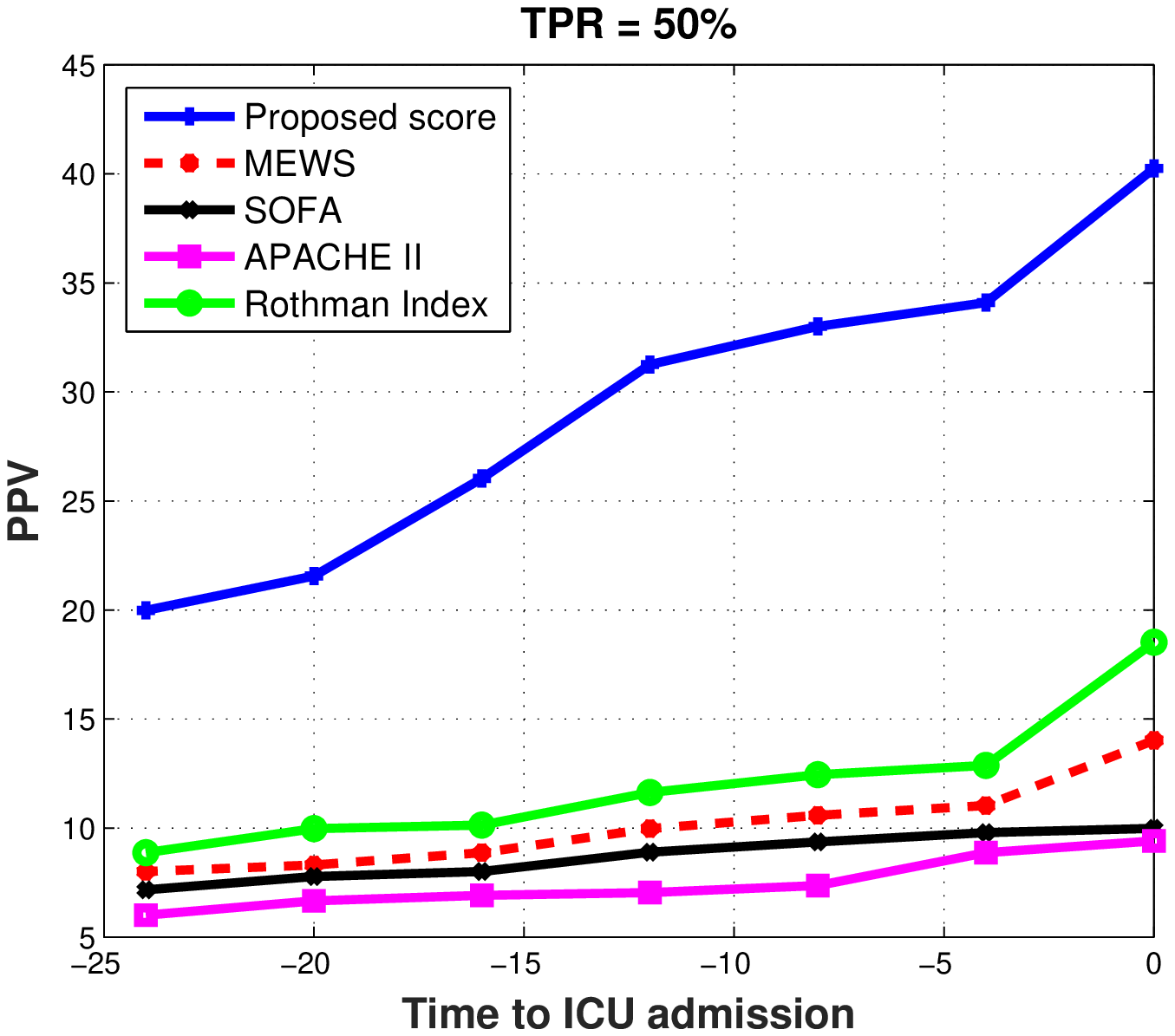}
        \caption{\small Timeliness curve.}		
	\label{Figa2}
  \end{minipage}
\end{figure}
We evaluated the prognostic utility of the proposed risk scoring algorithm by dividing the patient's cohort into a training set (admissions between March 2013 and November 2015) and a testing set (admissions between November 2015 and March 2016), estimating the model parameters from the training set, and then emulating the ICU admission decisions on the testing set by setting a threshold on the risk score $R_i(t)$, above which patient $i$ is identified as ``clinically deteriorating". The accuracy of such decisions are assessed via the following performance metrics: true positive rate (TPR), positive predictive value (PPV) and timeliness (i.e. the difference between the time of actual ICU admission as decided by clinicians and the time at which $R_i(t)$ exceeds the threshold). Using the Bayesian information criterion, we selected an instantiation of our model with 12 epochs and 4 phenotypes.

\begin{table}[t]
    \centering
    \caption{Performance comparisons for various machine learning algorithms ($p < 0.01$).}
\begin{tabular}{|c||c|c|c|c|c|c|c|c|c|c|c|c|}  
\toprule[2.5pt]
{\bf Algorithm} & \begin{turn}{90} Proposed risk score\end{turn} & \begin{turn}{90}ForecastICU\end{turn} & \begin{turn}{90}Random Forest\end{turn} & \begin{turn}{90}Logistic regression\end{turn} & \begin{turn}{90}LASSO\end{turn} & \begin{turn}{90}RNN\end{turn} & \begin{turn}{90}HMM\end{turn} & \begin{turn}{90}MTGP\end{turn} & \begin{turn}{90}Rothman\end{turn} & \begin{turn}{90}MEWS\end{turn} & \begin{turn}{90}SOFA\end{turn} & \begin{turn}{90}APACHE\end{turn} \\
\hline
\hline
{\bf AUC (ICU admission)}      & 0.45 & 0.39 & 0.36 & 0.27 & 0.26 & 0.29 & 0.32 & 0.3 & 0.25 & 0.18 & 0.13 & 0.14 \\ \hline
{\bf AUC (Discharge at 0.01)}  & 0.36 & 0.3 & 0.2  & 0.17 & 0.16 & 0.18 & 0.23 & 0.23 & 0.25 & 0.18 & 0.1  & 0.13 \\ \hline
{\bf AUC (Discharge at 0.05)}  & 0.34 & 0.25 & 0.17 & 0.15 & 0.13 & 0.17 & 0.19 & 0.2 & 0.24 & 0.18 & 0.1  & 0.13 \\ \hline
{\bf AUC (Discharge at 0.2)}   & 0.23 & 0.19 & 0.13 & 0.1  & 0.1  & 0.13 & 0.16 & 0.14 & 0.14 & 0.17 & 0.1  & 0.1 \\ \hline
\bottomrule[2.5pt]
\end{tabular}
\label{Tab2}
\end{table}

In Figure \ref{Figa1}, we compare the ROC curve of the proposed risk score with those of the Rothman index and MEWS scores. In addition, we compare our risk score with the APACHE II and SOFA scores; both scores were originally developed to predict mortality in the ICU but were recently shown to possess significant predictive power for predicting clinical deterioration in wards \cite{yu2014comparison}. As we can see in Figure \ref{Figa1}, the proposed risk score significantly outperforms all the other risk scores in terms of the AUC. In particular, the proposed risk score's AUC is 20$\%$ greater than that of the Rothman index, the best performing clinical risk score. Since over 200,000 in-hospital cardiac arrests occur in the U.S. annually \cite{merchant2011incidence}, those performance improvements gains correspond to thousands of lives saved each year. Moreover, as we can see in Figure \ref{Figa1}, the proposed risk score offers a greater value for the PPV at any TPR value. This result has a great implication in clinical practice; the proposed risk score can ensure more confidence in its issued ICU alarms, which would mitigate alarm fatigue and enhance a hospital's resource utilization \cite{subbe2001validation}. The key behind the performance gains achieved by our risk score is that it considers the patient's entire physiological trajectory and not just the current physiological measurements, thus it is not impulsively triggered by instantaneous physiological data that may not be truly reflective of clinical deterioration, thereby reducing the rate of false alarms. 

In Figure \ref{Figa2}, we fix the TPR value at 50$\%$ and evaluate the timeliness of various risk score at different values of the PPV. The resulting curve, which we call the {\it timeliness curve}, illustrates the trade-off between the timeliness of the ICU alarms and the false alarm rates, i.e. the more quick a risk score issues ICU alarms, the more likely it will exhibit a false alarm. As we can see in Figure \ref{Figa2}, the proposed risk score is always many hours ahead of all other scores for any value of the PPV, and can help predict ICU admissions many hours before a clinician would do at a reasonable PPV and TPR. For instance, the proposed risk score can predict ICU admissions 8 hours before clinicians for a PPV of 35$\%$ and TPR of 50$\%$. By issuing prompt alarms for clinically deteriorating patients, the proposed risk score with can provide clinicians with a safety net to focus their attentions on patients who are more likely to deteriorate many hours before they exhibit severe decompenstation, allowing for timely ICU admission and more efficient therapeutic interventions.     

We also compared our risk score with other machine learning algorithms, including random forests, ForecastICU, logistic regression, recurrent neural networks (RNNs), hidden Markov models with Gaussian emissions (HMMs), multi-task GP regression (MTGPs) \cite{ghassemi2015multivariate}, and LASSO. Each of these algorithms is trained using a window of physiological measurements that precedes the ICU admission or discharge decision, and the size of such window is optimized separately for every algorithm. The AUC results for all the algorithms under consideration are listed in Table \ref{Tab2}, and as we can see, the proposed outperforms all the competing algorithms, including our previous work in \cite{yoon2016forecasticu}, which did not consider individualization. We also evaluated the AUC of our risk score and all competing algorithms when jointly issuing both ICU and discharge alarms. That is, we set a lower risk threshold, below which the patient is discharged home, and an upper threshold above which the patient is transferred to the ICU. We computed the AUC of all risk scores when fixing the lower risk threshold at values 0.01, 0.05 and 0.2, and sweeping the upper risk threshold from 0 to 1. In all cases, the proposed risk score outperforms all the other benchmarks. Thus, the proposed risk score can help making better utilization for hospital resources by discharging patients who are clinically stable from the ward in a timely manner.      

\section{Conclusions}
In this study, a novel real-time risk score for actionable ICU prognostication is developed and validated. Unlike state-of-the-art risk models, the proposed risk score incorporates both the patients' non-stationary temporal physiological information and their individual baseline co-variates in order to accurately describe the patients' physiological trajectories. Experiments conducted on a cohort of 6,094 patients admitted to the Ronald Reagan UCLA medical center show that the proposed risk score is significantly more accurate than currently deployed risk scores and other machine learning algorithms.  
\bibliography{htp-amia} 

\end{document}